\documentclass{article}

\usepackage{PRIMEarxiv}

\usepackage[utf8]{inputenc} 
\usepackage[T1]{fontenc}    
\usepackage{hyperref}       
\usepackage{url}            
\usepackage{booktabs}       
\usepackage{amsfonts}       
\usepackage{nicefrac}       
\usepackage{microtype}      
\usepackage{lipsum}
\usepackage{fancyhdr}       
\usepackage{graphicx}       
\graphicspath{{media/}}     

\usepackage{amsmath}
\usepackage{amssymb}
\usepackage{amsfonts}
\usepackage{graphicx}
\usepackage[ruled,linesnumbered]{algorithm2e}
\usepackage{lineno,hyperref}
\usepackage{subfig}
\usepackage[table]{xcolor}
\usepackage{graphicx}
\usepackage[english]{babel}
\usepackage[utf8]{inputenc}
\usepackage[noend]{algpseudocode}
\usepackage{color,soul}
\usepackage{hyperref}
\modulolinenumbers[1]
\usepackage{lineno}
\usepackage{graphicx}
\title{M2GAN: A Multi-Stage Self-Attention Network for Image Rain Removal on Autonomous Vehicles
}

\author{
  Duc Manh Nguyen \\
  Vietnam National Space Center  \\
  Vietnam Academy of Science and Technology \\
  Hanoi, Vietnam\\
  \texttt{manh.nguyenduc.uet@gmail.com} \\
   \And
  Sang-Woong Lee \\
  Pattern Recognition and Machine Learning Lab \\
  Gachon University \\
  Seoul, South Korea\\
  \texttt{slee@gachon.ac.kr} \\
}

\begin{document}
\maketitle

\begin{abstract}
Image deraining is a new challenging problem in applications of autonomous vehicles. In a bad weather condition of heavy rainfall, raindrops, mainly hitting the vehicle's windshield, can significantly reduce observation ability even though the windshield wipers might be able to remove part of it. Moreover, rain flows spreading over the windshield can yield the physical effect of refraction, which seriously impede the sightline or undermine the machine learning system equipped in the vehicle.  In this paper, we propose a new multi-stage multi-task recurrent generative adversarial network (M2GAN) to deal with challenging problems of raindrops hitting the car's windshield. This method is also applicable for removing raindrops appearing on a glass window or lens. M2GAN is a multi-stage multi-task generative adversarial network that can utilize prior high-level information, such as semantic segmentation, to boost deraining performance. To demonstrate M2GAN, we introduce the first real-world dataset for rain removal on autonomous vehicles. The experimental results show that our proposed method is superior to other state-of-the-art approaches of deraining raindrops in respect of quantitative metrics and visual quality. M2GAN is considered the first method to deal with challenging problems of real-world rains under unconstrained environments such as autonomous vehicles.
\end{abstract}


\section{Introduction}

\begin{figure}
 \centering
    \includegraphics[width=0.9\linewidth]{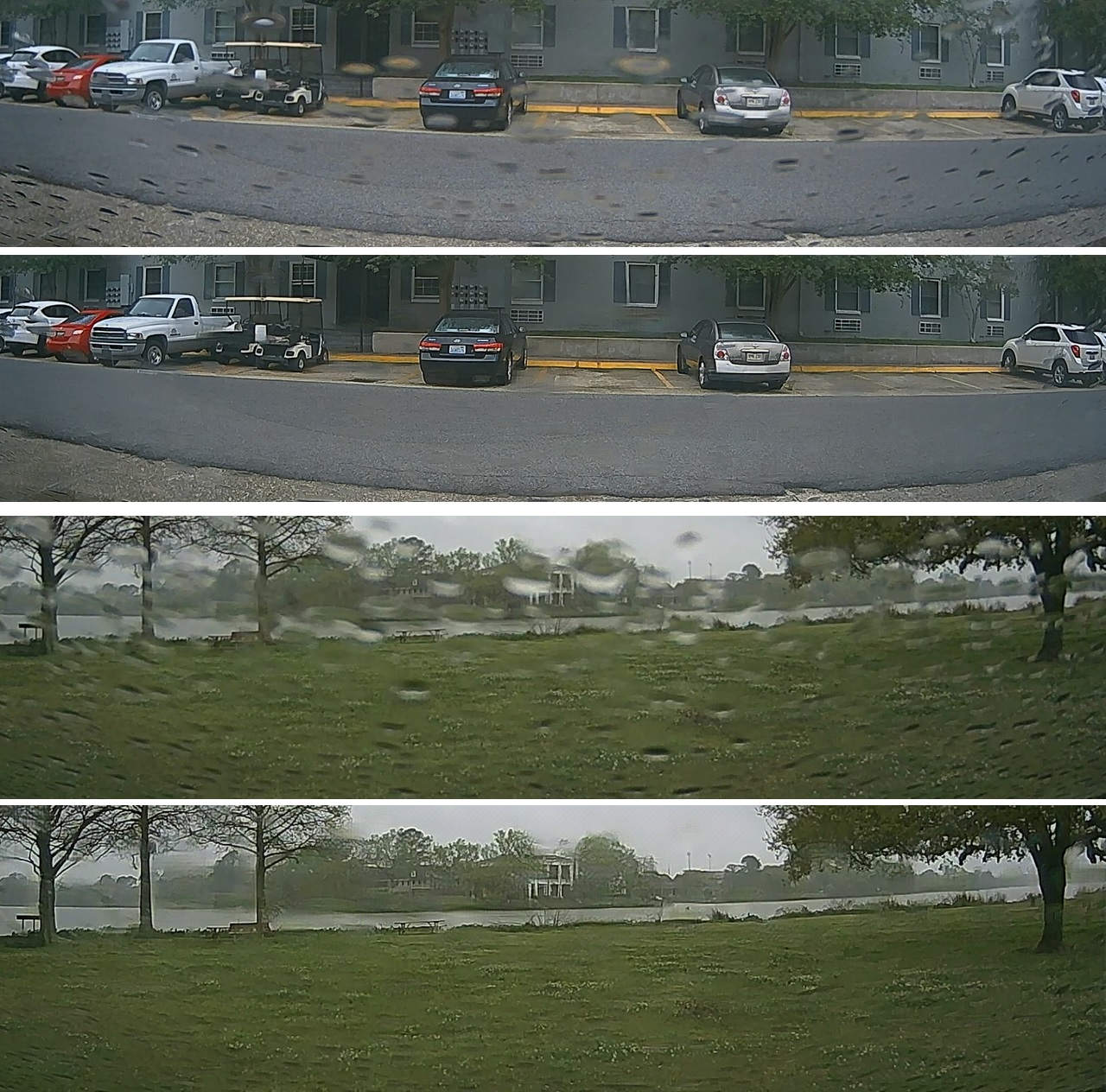} 
  \caption{Demonstration of the proposed method. Rows 1 and 3: Original images with real-world raindrops and rain flow on a car's windshield. Rows 2 and 4: Our results.}
  \label{fig_derain_1}
\end{figure}

Rain at different levels on an image severely reduces the performance of vision systems such as object detection and semantic segmentation of an autonomous vehicle \cite{SZang2019} \cite{GVolk2019} \cite{HKurihata2005} \cite{SHasirlioglu2018} \cite{VFischer2019} \cite{FBernardin2013}. Under severe weather such as heavy rain, the self-driving car is not able to drive safely and proficiently because camera systems become obscured by rain flows, while videos or images recorded from them can be significantly distorted. Although many approaches have proposed to deal with the adverse impacts of rainy weather, image deraining poses four challenging problems associated with the inherent nature of rain. 

First, image deraining is considered a challenging, ill-posed problem because rain, falling from the sky or hitting the windshield of a driver-less car, varies in a wide variety of shapes, scales, and densities, depending on the distances between raindrops and the camera. Moreover, rain streaks in heavy rain are highly unpredictable in directions and velocities and cause fuzzy scenes and blurry objects in the image \cite{TWang2019} \cite{YZheng2019} \cite{SDeng2019}. Additionally, the probability distributions of the rain streaks vary widely in different local regions and color channels in the image. 

Second, the most challenging problem in image deraining on autonomous vehicles is to recover context information corrupted by raindrops, hitting and spreading mainly on the surface of the windshield, whereas the camera system is mounted inside the car for safety.  Rain flows that spread over the windshield yield the physical effect of refraction \cite{JCHalimeh2016} \cite{SYou2013}, which seriously make self-driving cars not able to drive proficiently and detect objects accurately. The raindrop refraction is a physic phenomenon that happens when the sunlight passes through raindrops, subsequently changes its direction, and causes a severe illusion. This illusion makes the camera system cannot estimate accurately shapes as well as real positions of objects occurred in the image.  

\begin{figure}
 \centering
    \includegraphics[width=1\linewidth]{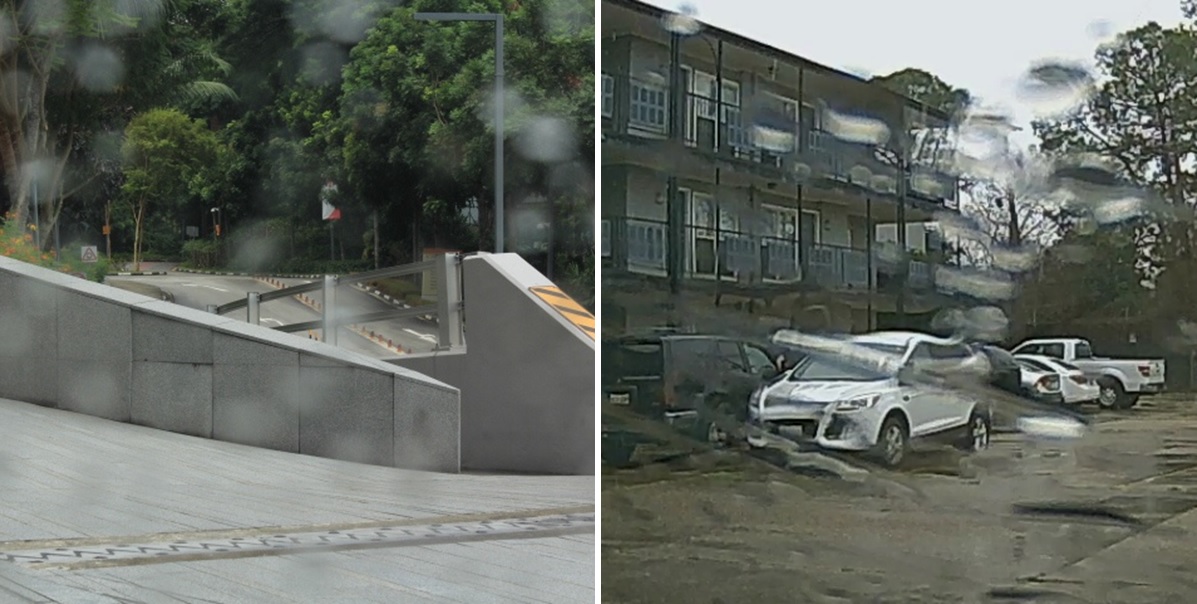} 
  \caption{Demonstration of the proposed method. Right: Image from our dataset, degraded by real-world raindrops and rain flows. Left: Image from the Raindrop dataset \cite{RQian2018}, degraded by waterdrops generated in controlled conditions.}
  \label{fig_derain_2}
\end{figure}

The third problem is the task-oriented deraining problem. A successful deraining method has to satisfy two main requirements at the same time. First, it is required to eliminate rain from the image background effectively. Second, it can reconstruct the whole scenarios in the image and guarantee that the corresponding cognitive system, such as object detection and semantic segmentation, performs well.

The final problem we aim to address in this paper is the lack of real-world training datasets for autonomous vehicles. There are several datasets of raindrops \cite{RQian2018} \cite{DEigen2013}, but their quality is limited and can not be applicable for autonomous vehicles applications. Their training images are collected by adding artificial raindrops or unreal rain streaks into clean background images. For instance, Qian et al. \cite{RQian2018} provided the Raindrop dataset in which rain images were generated under strictly controlled environments. The author also proposed an attentive generative adversarial network (AttentionGAN) trained on this dataset. In this dataset, the background image was corrupted by adding raindrops attached to a glass pane. Thus, the raindrops that appeared in the image were not real-world raindrops, and its physical model is far different from the real-world rain model. Particularly, raindrops in this dataset have similar shapes and sizes and are relatively easy to remove. As seen in Figure \ref{fig_derain_2}, the water drops attached in the images from the Raindrop dataset are consistently round, small, and thin. The features and details overlapped by the water drops are only partly blurred or slightly wiped out. Thus, these features are partly preserved and are not so difficult to recover from raindrops degradation. In contrast, the real-world raindrops will hit and spread mainly on the surface of the windshield of a car, causing unpredictable water flows as well as complex raindrops on the windshield. Moreover, as seen in Figure \ref{fig_derain_2}, the raindrops cause more severe effects of refraction than those in the Raindrop dataset, leading to a challenging task of recovering the original details of the image.

Motivated by addressing the problems of the physical rain effects and task-oriented deraining, we propose a new multi-task generative adversarial network, including a generator and two separate discriminators. The first discriminator is to classify derained and ground-truth images. The generator competes with the first discriminator to create plausible-looking natural images. The second discriminator aims to distinguish between semantic segmentation and the corresponding segmentation ground-truth maps. The second discriminator's key role is to guarantee that the generator can reconstruct reliable features and details for high-level recognition tasks and eliminate unreal artifacts deteriorating decision-making systems. As a result, the generator can yield the best-derained image and preserve the essential task-specific features for autonomous vehicles' applications, as shown in Figure \ref{fig_derain_1}. 

Since the distributions of raindrops and flows in front of windshields are complicated, the generator should be built based on a robust deep learning network. However, deeper networks might have a better performance, but they frequently face the gradient vanishing problem. Thus, we propose a multi-stage recurrent generative adversarial network that can eliminate different rain layers over stages. The quality of derained images can be improved over stages when raindrops and flows are removed gradually. To transfer information between generative networks at different stages, we propose using two kinds of attention maps: the attention rain map and the semantic segmentation map. The attention rain map helps the generator focus on local rain regions and better eliminate complex raindrops and flows. Since rain flows on the windshield can severely contaminate textures and details in the image, we use the semantic segmentation map as reliable prior information to improve deraining performance, as proved by Wang et al. \cite{XWang2018}. This map helps the generator classify local areas into specific categories. Hence, features and textures from the local areas can be restored reliably and truly instead of unreal artifacts.

To demonstrate our methods, we built up a reliable dataset with a large number of training pairs of real-world rain and ground-truth images. A rain image from our dataset might include raindrops and rain flows attached to a car's windscreen. Moreover, rain streaks exist in every rain image from our dataset even though its negative impact on the image is not as severe as the impacts caused by raindrops or rain flows. Rain streaks result in fuzzy scenes and blurry objects on the background. Rain streaks and rain flows did not exist in the other dataset of raindrops \cite{RQian2018} \cite{DEigen2013} despite their strong impacts on the results of reconstructing a rain image. State-of-the-art methods of removing raindrops performed well on synthesized datasets of rain and clean image pairs, which could not correctly characterize rainfall's nature in realistic environments. However, they failed when testing on our dataset of real-world rain images. 

We tested our proposed algorithms and competitors on our dataset of realistic rain images. In short, our main contributions are:

\begin{itemize}

\item We propose a multi-stage multi-task generative network to remove rain streaks and flows in the image effectively and recover realistic details and features deteriorated by rain.

\item We present the attention rain map and the semantic segmentation map that help the networks update and improve deraining performance.

\item We introduce the first real-world dataset for deraining tasks on autonomous vehicles. Since the windshield on a car is similar to a glass window or lens, our dataset is suitable for practical applications of eliminating raindrops under various uncontrolled environments.    

\end{itemize}

\section{Related Work}

\begin{figure*}
 \centering
    \includegraphics[width=0.8\linewidth]{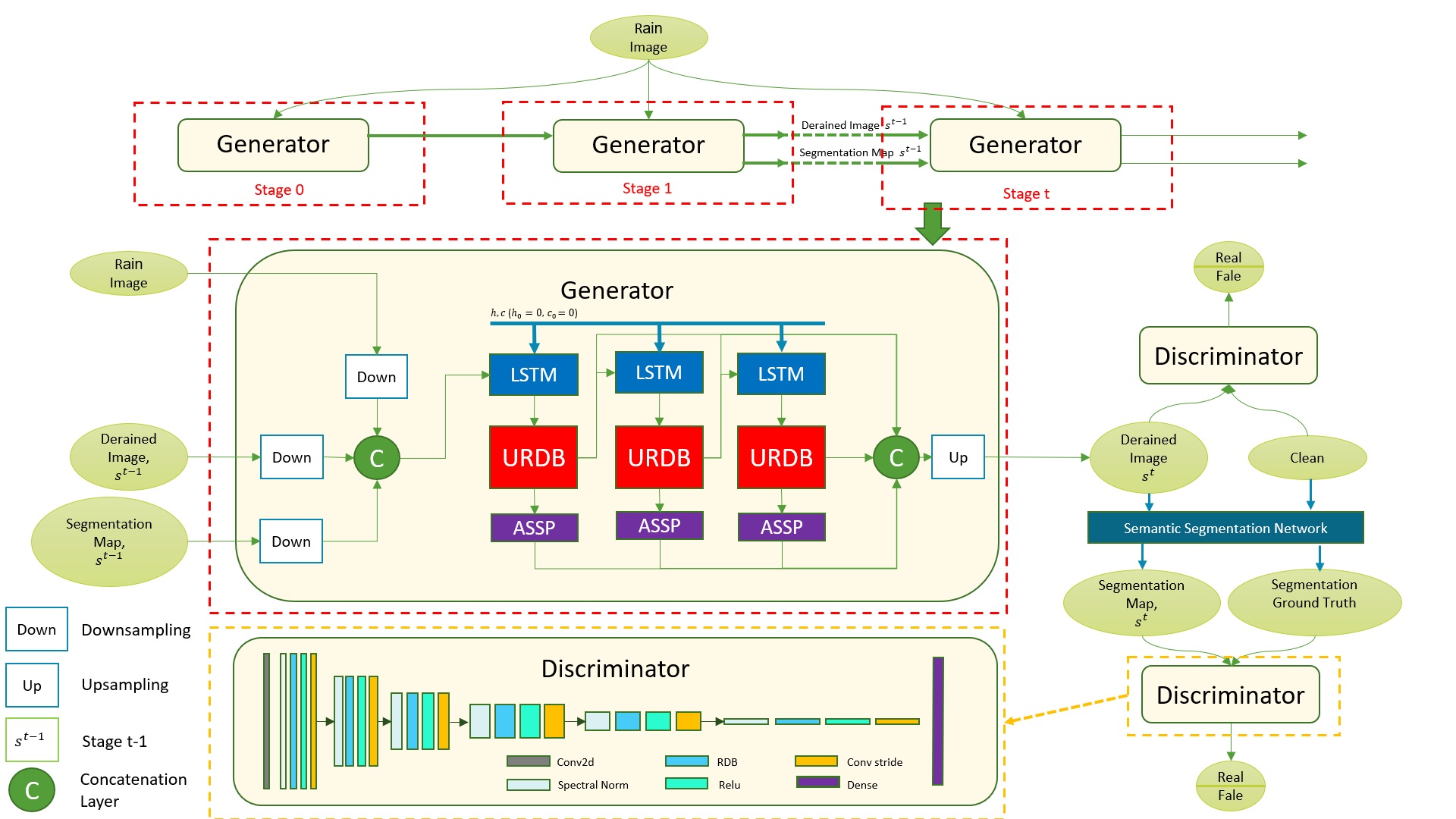} 
  \caption{Block diagram of the architecture of M2GAN.}
  \label{fig_resnet}
\end{figure*}

Recently, there has been more and more demand for eliminating rain streaks caused by bad weather \cite{XLi2018} \cite{TWang2019} \cite{YZheng2019} \cite{SDeng2019}. In order to address problems of raindrops, some methods \cite{RTTan2018} \cite{MRoser2009} \cite{MRoser2010} have been focused on detecting raindrops but cannot completely remove them. Kurihata et al. \cite{HKurihata2005} developed a statistical learning method to detect raindrops on windshields. However, this method can only remove tiny raindrops and generate many false positives and blurry outputs when the texture and the raindrop appearance are highly complex. This research also indicated that stationary drops strongly impact drivers' safety as well as the advanced driver-assistance system in a driverless car.  Nashashibi et al. \cite{Nashashibi2010} introduced the model of the sophisticated appearance of unfocused raindrops on the windshield. In this model, raindrops cannot be correctly separated from the scene background. Due to the physical phenomenon of refraction, objects seen by the camera system through unfocused raindrops turn out to be bigger, have rounded shapes, are difficult to recognize their contours, and have a severe blurring effect. You et al. \cite{SYou2013} \cite{SYou2016} utilized motion information to improve the performance of detecting and removing raindrops. This method, for this reason, fails to detect rain in a single image. Interestingly, Halimeh et al. \cite{JCHalimeh2016} presented a physical model of raindrop geometry on the windshield of a car. This model helps us understand the photometric properties of raindrops as well as the negative impacts of raindrops on vision-based driver assistance systems. Garg et al. \cite{KGarg2004} \cite{KGarg2007} demonstrated that the visual effects of rain in a nature image are derived from the appearance of a stationary drop, causing the problem of refraction, and rain streaks, leading to the popular issue of motion blur. Deep convolutional networks \cite{YZhang2018} \cite{YZheng2019} cannot effectively remove rain from natural rainy images such as those captured from an autonomous vehicle due to the problem of refraction. Interestingly, to deal with this intractable problem, Qian \cite{RQian2018} employed a generative adversarial network (GAN) to produce reconstructed images closing to natural images. This network is mainly formed by consecutive deep residual networks (ResNets) \cite{KHe2018} and convolutional LSTM layers \cite{Xingjian2015}.

\section{Methodology}

Given a rain image $\mathbf{O}\in \mathbb{R}^{M\times N}$, we train a set of generative networks in $\mathbf{N}_{s}$ stages. These networks are combined to estimate the background image $\mathbf{\hat{B}}\in \mathbb{R}^{M\times N}$, which is similar to its ground truth image $\mathbf{B}\in \mathbb{R}^{M\times N}$. At each stage $k$ with $k = 1,..,\mathbf{N}_{s}$, the generative network $\mathbf{G}_{k}$ is trained to directly map the rain image $\mathbf{O}$ to the estimated background image $\mathbf{\hat{B}}_{k}$ as follows

\begin{equation}
\mathbf{\hat{B}}_{k}=\mathbf{G}_{k}(\mathbf{O})
\end{equation}
 
We use the total loss function $\mathbf{L}_{G_{k}}^{total}$ for optimizing the performance
of $\mathbf{G}_{k}$ as 

\begin{equation}
\mathbf{\hat{\theta}}_{k}=argmin\sum_{i=1}^{\mathbf{N}_{tr}}\mathbf{L}_{G_{k}}^{total}(\mathbf{B},\mathbf{\hat{B}})
\end{equation}
where $\mathbf{N}_{tr}$ is the number of training image pairs, $\theta$ is training parameters of $\mathbf{G}_{k}$, and $\hat{\theta}$ is the optimal parameters of $\mathbf{G}_{k}$. $\mathbf{L}_{G_{k}}^{total}$ is the weighted sum of the perceptual loss $\mathbf{L}_{G_{k}}^{des}$ generated by the DenseNet network \cite{GHuang2017}, the adversarial loss $\mathbf{L}_{G_{k}}^{adv}$, and the mean absolute error (MAE) loss $\mathbf{L}_{G_{k}}^{mae}$. $\mathbf{L}_{G_{k}}^{total}$ is mainly designed to deal with the ill-posed problem of refraction  caused by raindrops hitting the wind-shield of the autonomous car. Thus, to circumvent the loss of the image textures and details, the perceptual loss $\mathbf{L}_{G_{k}}^{des}$, which is closer to perceptual similarity,  is used to optimize the performance of $\mathbf{G}_{k}$.  Moreover, the adversarial loss $\mathbf{L}_{G_{k}}^{adv}$ stimulates our network to focus on the solutions that lie on the manifold of natural images by attempting to fool the discriminator network. Finally, the MAE  loss is used to reduce the artifacts and noise.

To deal with the challenging problem of refraction, we aim to explore reliable prior information which is able to help generative networks restore natural objects and textures, such as buildings, roads, trees. In this paper, two kinds of prior information are used: the attention rain map and the semantic segmentation map. Thus, Eq. (3) can be modified as follows

\begin{equation}
\mathbf{\hat{B}}_{k}=\mathbf{G}_{k}(\mathbf{O}|\mathbf{M}_{k-1}^{R}, \mathbf{M}_{k-1}^{S})
\end{equation}
where $\mathbf{M}_{k-1}^{R}$ and $\mathbf{M}_{k-1}^{S}$ are the attention rain map and the semantic segmentation map at stage $k-1$, respectively. By using $\mathbf{M}_{k-1}^{R}$ and $\mathbf{M}_{k-1}^{S}$ as prior information, $\mathbf{G}_{k}$ can be considered a conditional GAN which can discriminate fake and real images better.

$\mathbf{M}_{k-1}^{R}$ can be directly generated from the output of $\mathbf{G}_{k-1}$ at stage $k-1$. $\mathbf{M}_{k-1}^{R}$ has two main advantages. First, $\mathbf{M}_{k-1}^{R}$ helps $\mathbf{G}_{k}$ focus on local rain regions and learn to eliminate complex rain streaks and raindrops better. Second, $\mathbf{M}_{k-1}^{R}$ plays a vital role in restoring and transferring information between generative networks at different stages. Deeper networks perform well, but they are frequently facing the gradient vanishing problem.  $\mathbf{M}_{k-1}^{R}$ helps a multi-stage network work better than a single deep network. 

\begin{figure}
 \centering
    \includegraphics[width=0.7\linewidth]{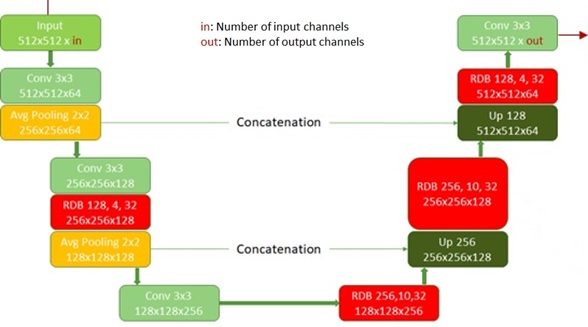} 
  \caption{Block diagram of the architecture of URDB.}
  \label{fig_URDB}
\end{figure}

$\mathbf{M}_{k-1}^{S}$ is an output of a semantic segmentation network with the same input image with $\mathbf{G}_{k-1}$ at stage $k-1$. We aim to train a multi-stage multi-task generative adversarial network to improve both results of deraining and semantic segmentation. Since the semantic segmentation performance can be effectively improved at the first stage, the semantic segmentation map becomes robust prior information to enhance the quality of the derained image. Our network is illustrated in Figure \ref{fig_resnet}.

\subsection{Multi-Task Generative Adversarial Networks}

\begin{figure*}
 \centering
    \includegraphics[width=1\linewidth]{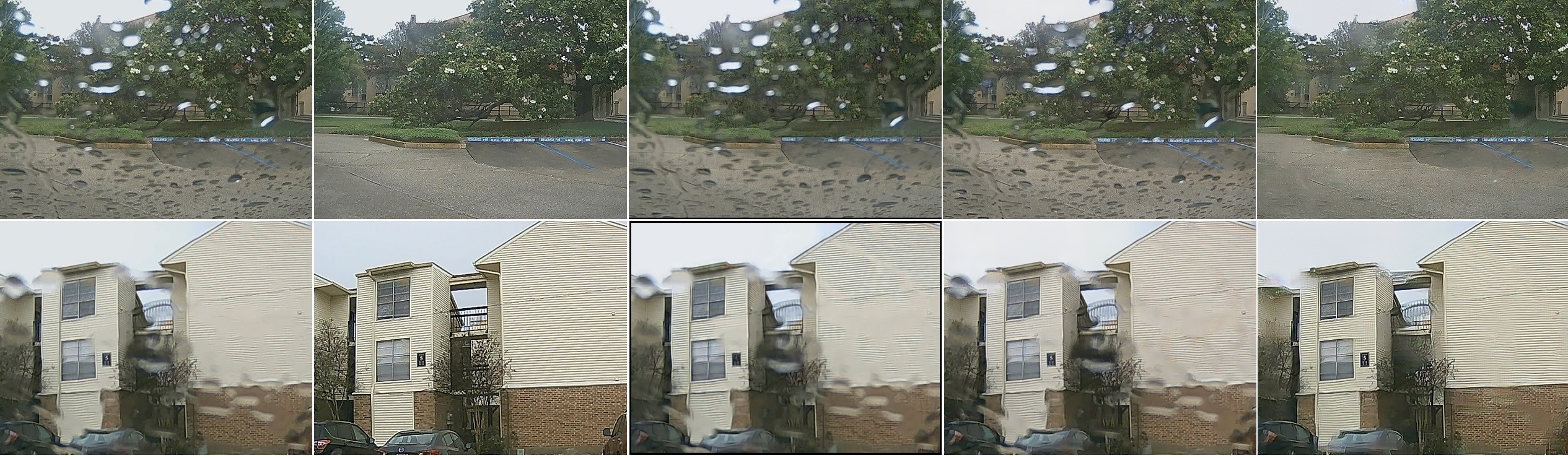} 
  \caption{Visual comparisons of M2GAN and its competing methods on our dataset. Column 1: Rain images. Column 2: The corresponding ground-truth images. Column 3: The results of Eigen13 method. Column 4: The results of AttentionGAN network. Column 5: The results of M2GAN.}
  \label{fig_derain_3}
\end{figure*}

Recently, most of the existing deraining methods only focus on increasing the peak signal-to-noise ratio (PSNR). However, this criterion is very limited in evaluating the ability to preserve object details and structures. The use of PSNR does not sufficiently assess the deraining performance in the study of autonomous vehicles. In the study of robots or autonomous cars, a successful deraining method has to satisfy two main requirements at the same time. First, it is required to eliminate rain from the image background effectively. Second, it can reconstruct the whole scenarios in the image and guarantee that high-level machine learning-based systems, such as object detection and semantic segmentation, perform well. Recent advanced GAN-based methods \cite{Vo2021a} \cite{Vo2021b} have an effective strategy to increase the accuracy of the discriminator. For this reason, we present a new multi-task adversarial network, including a generator and two separate discriminators. One discriminator is to classify derained and ground-truth images. The other one aims to distinguish between the semantic segmentation map and the corresponding segmentation ground-truth map. During training, the generator and discriminator sub-networks compete with each other. As a result, the generator can yield the best-derained image and preserve the essential task-specific features for high-level object recognition and semantic segmentation. 

Specifically, at stage $k$, the generative network $\mathbf{G}_{k}$ competes with the semantic segmentation discriminator $\mathbf{D}_{k}^{s}$ and the deraining discriminator $\mathbf{D}_{k}^{d}$. $\mathbf{G}_{k}$ and $\mathbf{D}_{k}^{d}$ are trained together to generate derained images that can preserve more textures and edges while also having a high PSNR value. Jolicoeur-Martineau (2018) demonstrated that GAN-based methods using a relativistic discriminator are more stable than the standard one. Thus, we use the formula $\mathbf{D}_{k}^{d}(\tilde{x})=sigmoid(\mathbf{\Phi_{k}}(x_{r})-\mathbf{\Phi_{k} }(x_{f}))$ to make the discriminator $\mathbf{D}_{k}^{d}$ relativistic and more stable, which $\mathbf{\Phi_{k} }(x)$ is the output non-transformed layer of $\mathbf{D}_{k}^{d}$. In this formula, $\tilde{x}=(x_{r},x_{f})$ is a pair of real and fake images. In fact, $\mathbf{D}_{k}^{d}(\tilde{x})$ is the probability that the fake image $x_{f}$ is more realistic than its corresponding ground-truth image. Similarly, we use the formula $\mathbf{D}_{k}^{s}(\tilde{z})=sigmoid(\mathbf{\Omega_{k}}(z_{r})-\mathbf{\Omega_{k} }(z_{f}))$ to train the discriminator $\mathbf{D}_{k}^{s}$, which $\mathbf{\Omega_{k} }(z)$ is the output non-transformed layer of $\mathbf{D}_{k}^{s}$. In this formula, $\tilde{z}=(z_{r},z_{f})$ is a pair of real and fake segmentation maps extracted from an external segmentation network in which the inputs are $x_{r}$ and $x_{f}$, respectively. Particularly, the adversarial loss functions of $\mathbf{D}_{k}^{d}$, $\mathbf{D}_{k}^{s}$, and $\mathbf{G}_{k}$ can be computed by the following formulas, respectively:

\begin{equation}
\begin{aligned}
\mathbf{L}_{\mathbf{D}_{k}^{d}}(x_{f}, x_{r})=&\mathbf{E}_{x_{r}\sim \mathbb{P}}\begin{bmatrix}
\mathbf{f}_{1}(\mathbf{C}(x_{r})-\mathbf{E}_{x_{f}\sim \mathbb{Q}}\mathbf{C}(x_{f}))
\end{bmatrix} \\
+&\mathbf{E}_{x_{f}\sim \mathbb{Q}}\begin{bmatrix}
\mathbf{f}_{2}(\mathbf{C}(x_{f})-\mathbf{E}_{x_{r}\sim \mathbb{P}}\mathbf{C}(x_{r}))
\end{bmatrix}
\end{aligned}
\end{equation}

\begin{equation}
\begin{aligned}
\mathbf{L}_{\mathbf{D}_{k}^{s}}(z_{f}, z_{r})=&\mathbf{E}_{z_{r}\sim \mathbb{U}}\begin{bmatrix}
\mathbf{f}_{1}(\mathbf{C}(z_{r})-\mathbf{E}_{z_{f}\sim \mathbb{V}}\mathbf{C}(z_{f}))
\end{bmatrix} \\
+&\mathbf{E}_{z_{f}\sim \mathbb{V}}\begin{bmatrix}
\mathbf{f}_{2}(\mathbf{C}(z_{f})-\mathbf{E}_{z_{r}\sim \mathbb{U}}\mathbf{C}(z_{r}))
\end{bmatrix}
\end{aligned}
\end{equation}

\begin{equation}
\begin{aligned}
\mathbf{L}_{G_{k}^{adv}}(x_{f}, x_{r})=&\mathbf{E}_{x_{r}\sim \mathbb{P}}\begin{bmatrix}
\mathbf{g}_{1}(\mathbf{C}(x_{r})-\mathbf{E}_{x_{f}\sim \mathbb{Q}}\mathbf{C}(x_{f}))
\end{bmatrix} \\
+&\mathbf{E}_{x_{f}\sim \mathbb{Q}}\begin{bmatrix}
\mathbf{g}_{2}(\mathbf{C}(x_{f})-\mathbf{E}_{x_{r}\sim \mathbb{P}}\mathbf{C}(x_{r}))
\end{bmatrix} \\
+&\mathbf{E}_{z_{r}\sim \mathbb{U}}\begin{bmatrix}
\mathbf{g}_{1}(\mathbf{C}(z_{r})-\mathbf{E}_{z_{f}\sim \mathbb{V}}\mathbf{C}(z_{f}))
\end{bmatrix} \\
+&\mathbf{E}_{z_{f}\sim \mathbb{V}}\begin{bmatrix}
\mathbf{g}_{2}(\mathbf{C}(z_{f})-\mathbf{E}_{z_{r}\sim \mathbb{U}}\mathbf{C}(z_{r})) \\
\end{bmatrix} \\
\end{aligned}
\end{equation}
where $\mathbf{f}_{1}(y)=\mathbf{f}_{2}(y)=-\mathbf{g}_{1}(y)=-\mathbf{g}_{2}(y)=sigmoid(y)$, $\mathbb{P}$ is the distribution of real clean images, $\mathbb{Q}$ is the distribution of fake derained images, $\mathbb{U}$ is the distribution of real segmentation maps, $\mathbb{V}$ is the distribution of fake segmentation maps. To increase PSNR and adaptively preserve important details in the original image, the MAE loss $\mathbf{L}_{G_{k}}^{MAE}(x_{f}, x_{r})$ and the DenseNet loss $\mathbf{L}^{des}_{G_{k}}(x_{f}, x_{r})$ are also computed. The calculation of the pixel-wise MAE loss is presented by:

\begin{equation}
\mathbf{L}_{G_{k}}^{MAE}(x_{f}, x_{r})=\left \| x_{r}-x_{f} \right \|_{1}^{1} 
\end{equation}

We also use the DenseNet loss function for optimizing the performance of the generator that is close to perceptual similarity. In this paper, we built the DenseNet network with four dense blocks with 162 convolution layers as described by Huang et al. \cite{GHuang2017}. The features are extracted only from the feature map of the third dense block in which low-level and high-level features are accumulated properly. Given an image pair $(x_{f}$, $x_{r})$, the computing of the DenseNet loss function is presented by:

\begin{equation}
\mathbf{L}^{des}_{G_{F}}(x_{f}, x_{r})=\frac{1}{K}\sum_{i=1}^{W_{u}}\sum_{j=1}^{H_{u}}\sum_{k=1}^{C_{u}}(\mathbf{\Psi} (x_{f})_{ijk}-\mathbf{\Psi} (x_{r})_{ijk})^{2}
\end{equation}
where $K=W_{u}H_{u}C_{u}$. Additionally, $W_{u}$, $H_{u}$ and $C_{u}$ denote the dimensions of the feature map $\Psi$ extracted from the DenseNet network. Finally, the weighted loss $\mathbf{L}_{G_{k}}^{total}$ in Eq. (4) can be modified by 

\begin{equation}
\mathbf{L}_{G_{k}}^{total}=\mathbf{\omega} _{1}\mathbf{L}_{G_{k}}^{MAE} + \mathbf{\omega} _{2}\mathbf{L}^{des}_{G_{k}} + \mathbf{\omega} _{3}\mathbf{L}_{G_{k}^{adv}} 
\end{equation}
which $\mathbf{\omega} _{1}$, $\mathbf{\omega} _{1}$, $\mathbf{\omega} _{3}$ are the regularization parameters. In this paper, we set $\mathbf{\omega} _{1}=0.1$, $\mathbf{\omega} _{2}=1$, and  $\mathbf{\omega} _{3}=0.001$

\section{Model Architecture}

\subsection{Model Architecture of The Generators}

\begin{figure*}
 \centering
    \includegraphics[width=1\linewidth]{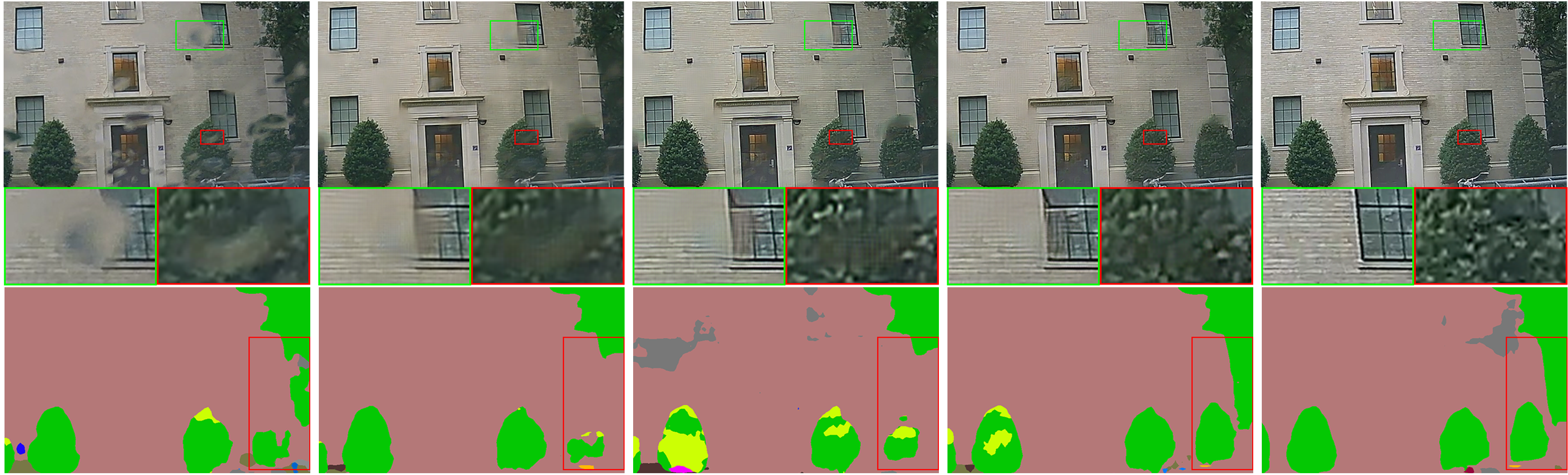} 
  \caption{Visual comparisons of different deraining methods on our dataset. Rows 1 and 2 shows RGB images and the corresponding semantic segmentation maps, respectively. Column 1: A rain image. Column 2: The corresponding ground-truth image. Column 3: The results of No-Disc network. Column 4: The results of No-Seg network. Column 5: The results of M2GAN. }
  \label{fig_derain_4}
\end{figure*}

Since deeper networks have a better performance, the generative network $\mathbf{G}_{k}$ consists of three consecutive URDB networks, as shown in Figure \ref{fig_resnet}. Each URDB combines the advantages of residual dense blocks (RDB) \cite{YZhang2018} and UNets \cite{Ronneberger2015}, which allows the maximum information flow to pass through all convolutional layers in the network and can probe hierarchical features for image restoration effectively. As seen in Figure \ref{fig_URDB}, URDB is a convolutional auto-encoder comprised of an encoding part to produce hierarchical high-level representative features and a decoding part to reconstruct the noisy image on context information distilled from the layers of its symmetry encoding part. Both encoding and decoding parts include multiple levels of which details and features can be extracted in different scales, using global average pooling and up-sampling layers.  The encoding part includes two contraction blocks. However, our contraction block is a modified residual dense block (RDB) which can utilize context information from the preceding contraction blocks, and fully extract sophisticated local features from all the layers within it, based on dense local connections. It is worth noting that the RDB of each contraction block has twice the number of kernels than that of the preceding contraction block so that URDB can effectively extract high-level features. The decoding part is also called the expansion part that consists of two expansion blocks. In order to maintain the symmetric property of URDB, each expansion block has only half of the number of kernels compared to its preceding expansion block. The output feature map of the last expansion block keeps going through another 3$\times$3 convolution layer to generate the reconstructed image. The inputs of each URDB are connected to an convolutional LSTM (ConvLSTM) \cite{Xingjian2015} layer's outputs to receive the input data and prior information on local rain areas and semantic segmentation classes. The output feature map of each URDB is connected to an ASPP layer to increase the whole network's receptive field and extract multi-scale features of the image.

\subsection{Discriminator}

The discriminator, $\mathbf{D}_{F}$, is also designed based on the architecture of RDBs, as shown in Figure \ref{fig_resnet}. Since RDBs are able to adaptively preserve all low-level and high-level features, $\mathbf{D}_{F}$ can be used for maximally discriminating between two probability distributions of fake and real images. Moreover, for stabilizing the training of $\mathbf{D}_{F}$, we utilize the novel weight normalization method that is termed spectral normalization \cite{Miyato2018}.

\section{A Real-world Rain Dataset for Autonomous Vehicles}

Recently, there has been no real-world dataset for deraining tasks on autonomous cars. Hence, we built up our dataset with a large number of training pairs of real-world rain and ground-truth images. To obtain real-world rain data, we used two different cars equipped with six surveillance cameras, which can quickly record a video of high-resolution 1920$\times$1080 training images. Ground-truth images had been collected before the rain started while the corresponding rain images were obtained when it was raining.  To have high-quality training images, we had to capture images when the background objects and the car were immobile. Thus, each pair of training images has the same static background captured in various street scenes and rain flows. In our dataset of real rain images, raindrops are highly random and diverse; however, the number of background scenes are limited due to the challenging problem of collecting real-world rain. To diversify the background images, we collected more synthetic rain images for training and testing. To capture synthetic rain images, we sprayed water in the air to simulate real raindrops, which can hit the windshield at a similar speed of actual raindrops, with random directions. As a result, water drops hitting the windshield have a similar distribution with the real-world raindrops. In total, we obtained 15,000 pairs of 720$\times$480 training images of real-world rain and 5,000 pairs of 720$\times$480 training images of synthetic rain. We collected other 400 pairs of 720$\times$480 testing images.

\section{Experimental Results}

\subsection{Training Settings}

Training patches with a randomly cropped size of 512$\times$512 were from the randomly sampled images from these datasets.  Based on the algorithm of Lookahead Optimizer \cite{RZhang2019}, which was used for minimizing cost functions, all our networks were trained effectively and can achieve faster convergence. The learning rate of the trained networks was set originally to 1e-3 and progressively reduced to 1e-5. We set the weight decay of each convolution layer to zero. Finally, the number of epochs was set to 50 for each network. All the segmentation map were extracted from the Resnet-101 convolutional neural network \cite{TXiao2018} \cite{Chen2017} trained on the ADE20K dataset \cite{BZhou2019}. Our models were trained by using the Pytorch open-source machine learning library \cite{paszke2017automatic}. We evaluated our methods on a Nvidia GTX 1080 Ti graphics card.

\subsection{Quality Measures}

We evaluate our method and its competing methods under autonomous car environments by using three commonly used metrics, which are PSNR, SSIM and FID \cite{Heusel2019}. To evaluate the performance of GAN-based methods mentioned in this paper, we also used the Fréchet Inception Distance (FID) score to measure the similarity of derained images to the corresponding ground-truth ones. Lower FID values mean that the GAN-based method generates more realistic images with less unreal artifacts. We did experiments on the rain dataset to compare the performance between our method and its the competing algorithms including Eigen13 \cite{DEigen2013} and AttentionGAN \cite{RQian2018}.

\subsection{Quantitative Evaluation}

Table \ref{table_Rain} indicates that M2GAN consistently outperforms all the competing methods. The PSNR and SSIM values by M2GAN are at least 1.97 dB and 0.0755 higher than those by its competitors, respectively. To evaluate the contributions of the main parts of our network, we also compare the whole system with those without main parts. No-Seg denotes the network without the segmentation discriminator. No-Disc denotes the network without both discriminators. As shown in Table \ref{table_Rain}, M2GAN outperforms the other networks, showing that the discriminators play crucial roles in preserving important details and preventing artifacts in the image. The FID score of M2GAN is at least 16.7334 lower than that of the other competing methods in this paper. It indicates that our multi-task GAN-based method is able to generate plausible-looking natural images with realistic details and features. 

We also demonstrate our method on the Raindrop dataset and show the results in Table \ref{table_Rain2}. Table \ref{table_Rain2} shows that our approach yielded higher PSNR than AttentionGAN \cite{RQian2018}. This proves that M2GAN is more effective in removing rain, unreal artifacts.

\begin{table}
\centering
\renewcommand{\arraystretch}{1.5}
\caption{The quantitative results on the autonomous vehicles dataset.}
\label{table_Rain}
\centering
\begin{tabular}{|p{2.5cm}<{\centering}|p{1.5cm}<{\centering}|p{1.2cm}<{\centering}|p{1.4cm}<{\centering}|p{1.2cm}}
\hline
Methods & FID   & PSNR & SSIM 
\\\hline\hline
\hline 
Eigen13 \cite{DEigen2013}  & 99.3800 & 17.84 & 0.6149
\\\hline
Qian \cite{RQian2018} & 46.4451 & 20.68 & 0.6620
\\\hline
No-Disc & 41.0400 & 21.51 & 0.6894
 \\\hline
No-Seg  & 37.1088 & 21.67 & 0.7146
\\\hline
M2GAN & $\mathbf{29.7117}$ & $\mathbf{22.65}$ & $\mathbf{0.7375}$
\\\hline
\end{tabular}
\label{table_Rain}
\end{table}

\begin{table}
\centering
\renewcommand{\arraystretch}{1.5}
\caption{The quantitative results on the Raindrop dataset \cite{RQian2018}.}
\label{table_Rain2}
\centering
\begin{tabular}{|p{0.9cm}<{\centering}|p{1.0cm}<{\centering}|p{1cm}<{\centering}|p{1.0cm}<{\centering}|p{1.0cm}<{\centering}|p{0.9cm}<{\centering}|p{0.9cm}}
\hline
Metric & Eigen13 \cite{DEigen2013} &  Isola (2016) & Qian \cite{RQian2018} & Quan (2019) & Ours 
\\\hline\hline
PSNR & 28.59 & 30.14 & 31.51 & 31.44 & $\mathbf{31.57}$
\\\hline
SSIM & 0.6726 & 0.8299 & 0.9213 & $\mathbf{0.9263}$ & 0.9157
\\\hline
\end{tabular}
\label{table_Rain2}
\end{table}

\subsection{Qualitative Evaluation}


Figure \ref{fig_derain_3} shows that M2GAN is the best method in balancing between erasing different kinds of raindrops and recovering texture details. Compared to our approach, Eigen13 and AttentionGAN fail to remove raindrops even with simple and small ones. Figure \ref{fig_derain_3} explains the failure of Eigen13 and AttentionGAN explicitly. These methods were developed to address the problem of water drops under strictly controlled environments. The Raindrop dataset was built by collecting pairs of degraded and clean images. All the degraded photos were captured through the same glass pane on which the author sprayed water drops randomly. The corresponding clean photos were captured by using another glass pane without spraying water. Based on this experiment, the raindrops were simulated under controlled environments. Consequently, they were significantly different from real-world rains in terms of distributions and physical effects. As seen in Figure \ref{fig_derain_1}, the water drops attached in the images from the Raindrop dataset are consistently round, tiny, and thin. As a result, the features and details overlapped by the water drops are blurred and slightly refracted. Thus, it is not difficult to recover the degraded features and details in the image. Similarly, the Eigen13 network was trained on the rain pictures taken through a glass pane on which some water drops were sprayed. The water drops attached to the glass pane are very small compared to the objects on the background. Consequently, the Eigen13 network inefficiently performs in the rain images from our challenging dataset in which the raindrops and rain flows have a wide variety of shapes and sizes. Unlike the Raindrop and Eigen13 datasets, our real-world rain dataset was developed under uncontrolled environments. All rain images in this dataset were captured in realistic rain conditions, reflecting the real distribution of raindrops and rain flows. As shown in Figure \ref{fig_derain_1}, the raindrops that appeared in the images are highly different from those from the Raindrop dataset. They have a wide variety of shapes, thickness, flow directions. The textures overlapped by the rain are severely degraded and refracted, which are very challenging to preserve. It explains the poor results of Eigen13 and AttentionGAN on our testing dataset. In contrast, our method provides promising performance leading to useful applications in practical conditions.

Figure \ref{fig_derain_3} shows the clear evidence that 
M2GAN shows the best performance in both semantic segmentation and deraining. No-Disc tends to smooth out the image details and textures and generate blurry local regions. No-Seg is able to preserve textures while removing rain from the background. However, No-Seg generates more unreal artefacts in the image than M2GAN. M2GAN is more effective than No-Seg and No-Disc in removing rain, unreal artefacts and preserving important details of the background, owing to the competition between the generator and the two discriminators.

\section{Conclusion}

Inspired by the idea of state-of-the-art deep learning-based methods, we proposed M2GAN to deal with challenging problems of raindrops hitting car's windshields. Extensive experimental results demonstrated that M2GAN performed considerably better than state-of-the-art methods in handling real-world raindrops and rain flows. In this paper, we demonstrate several vital contributions to solving the recent problems of deraining. First, we developed a framework of multi-stage generative adversarial networks to boost deraining performance. Second,we introduced the first real-world dataset for deraining raindrops. M2GAN is considered the first method that can significantly address the challenging problems of real-world raindrops under unconstrained environments.

\end{document}